\documentclass[preprint,12pt]{elsarticle}

\usepackage{amssymb}
\usepackage{subcaption}
\usepackage{amsmath,amssymb,amsfonts}
\usepackage{algorithmic}
\usepackage{graphicx}
\usepackage{textcomp}
\usepackage{multirow,multicol}
\usepackage{hyperref}
\usepackage{breqn}
\usepackage{colortbl}

\usepackage{adjustbox}
\usepackage{float}
\usepackage{makecell}
\setcellgapes{5pt}
\usepackage[table]{xcolor}
\usepackage{rotating}
\usepackage{pgfplots,tikz}

\usepackage[table]{xcolor}
\usepackage{nicematrix}

\pgfplotsset{compat=1.18}
\pgfplotscreateplotcyclelist{fc}{
fill=brown!80,draw=black,solid,mark=none\\
  fill=orange!80,draw=black,solid,mark=none\\
  fill=yellow!80,draw=black,solid,mark=none\\
   fill=red!80,draw=black,solid,mark=none\\
}

\newcolumntype{E}{>{\collectcell\ColCell}m{0.8cm}<{\endcollectcell}}

\newcommand\ColCell[1]{
        \ifnum#1=0
            \edef\x{\noexpand\centering\noexpand\cellcolor[RGB]{255,255,255}}\x #1
        \else
            \edef\x{\noexpand\centering\noexpand\cellcolor[RGB]{255,100,100}}\x #1
        \fi
    }

\journal{arXiv}

\begin{document}

\begin{frontmatter}

\title{Context-Aware Quantitative Risk Assessment Machine Learning Model for Drivers Distraction}

\author[inst1]{Adebamigbe Fasanmade}\ead{alex.fasanmade@dmu.ac.uk}
\author[inst1]{Ali H. Al-Bayatti}\ead{alihmohd@dmu.ac.uk}
\author[inst1]{Jarrad Neil Morden}\ead{jarrad.n.morden@dmu.ac.uk}

\affiliation[inst1]{organization={School of Computer Science and Informatics, De Montfort University},
            addressline={Tthe Gateway}, 
            city={Leicester},
            postcode={LE1 9BH}, 
            state={England},
            country={UK}}

\author[inst2]{Fabio Caraffini\corref{cor}}
\ead{fabio.caraffini@swansea.ac.uk}
\cortext[cor]{Corresponding author}

\affiliation[inst2]{organization={Department of Computer Science, Swansea University},
            addressline={Bay Campus, Computational Foundry, Fabian Way}, 
            city={Swansea},
            postcode={SA1 8EN}, 
            state={Wales},
            country={UK}}
\begin{abstract}

Risk mitigation techniques are critical to avoiding accidents associated with driving behaviour. We provide a novel Multi-Class Driver Distraction Risk Assessment (MDDRA) model that considers the vehicle, driver, and environmental data during a journey. MDDRA categorises the driver on a risk matrix as safe, careless, or dangerous. It offers flexibility in adjusting the parameters and weights to consider each event on a specific severity level. We collect real-world data using the Field Operation Test (TeleFOT), covering drivers using the same routes in the East Midlands, United Kingdom (UK). The results show that reducing road accidents caused by driver distraction is possible. We also study the correlation between distraction (driver, vehicle, and environment) and the classification severity based on a continuous distraction severity score. Furthermore, we apply machine learning techniques to classify and predict driver distraction according to severity levels to aid the transition of control from the driver to the vehicle (vehicle takeover) when a situation is deemed risky. The Ensemble Bagged Trees algorithm performed best, with an accuracy of 96.2\%.
\end{abstract}

\begin{keyword}
Risk assessment \sep Decision support \sep Driver behaviour \sep Severity rating \sep Driver distraction
\end{keyword}

\end{frontmatter}

\section{Introduction}\label{sec:introduction}

Since the emergence of new vehicle technologies, distracted drivers have become the main problem in road accidents. Meanwhile, intelligent transportation systems are getting closer to allowing vehicles to take control to a semiautonomous level 4 (or the level approved by the authorities) out of either necessity or driver choice. Thus, drivers may become more dependent on having the vehicle perform in-vehicle tasks, meaning they will become more relaxed and distracted, thus opening up many risks. In light of this, relevant contextual information (including vehicle performance and environmental conditions, which directly affect driver safety) can be used to help the driver cope with many situations. This implies the need for an Advanced Driving Assistance System (ADAS) to mitigate risks before an accident occurs by providing a qualitative- and quantitative-based risk assessment.

 The European Commission for Mobility and Road Transport Safety highlights that a significant proportion of road accidents occur when the driver is distracted, with common distractions encompassing handheld mobile devices, using the radio, eating, talking to passengers, smoking, and glancing at in-vehicle navigation systems \cite{b1}. 
 According to \cite{b2}, in-vehicle interfaces can also overload the driver. Additionally, fatigued drivers pose a significant risk on the road. In recent years, driver's eyes have become an efficient metric for measuring driver distraction, and driver's ability to keep their eyes on the road is crucial. A statistical analysis by the Department of Transport shows that of the $1,456$ fatal car accidents, $383$ involved careless tendencies of pedestrians, while 110 were the result of reduced attention of drivers on the road \cite{b3}. Inexperienced drivers are another significant factor that has caused the number of road accidents to increase. Young and inexperienced drivers are particularly at risk, unlike skilled drivers, who adjust their driving strategy in time and predict different driving scenarios \cite{b4}. In comparison, the higher incidence of accidents among young drivers is attributed to low cognitive ability \cite{b5} and a loss of attention due to distraction \cite{b6}.
Consequently, there is a strong need for driver risk assessment \cite{b7} which can provide an easy control change to automatic driving, especially in cases where the driver is intoxicated, unconscious, or not available in the situation. Although risk mitigation is tricky and difficult to model as official accident reports are relatively undetermined due to the possibility of numerous definitions of distractions or a country simply not collecting the data \cite{b8}. Additionally, driver distraction can be influenced by the situation in which driving occurs. Therefore, there is a significant gap in the available mechanism that accommodates the context-aware risk model. The model should be sufficiently concise with the ability of intelligent image recognition to detect and form a risk matrix to profile drivers into distraction classifications. This can reduce the occurrence of an accident by a significant margin.
There is a wide range of literature \cite{b9,b10,b11,b12,b13,b14} on the importance and urgency of such driving risk mitigation techniques to prevent driving behaviour-related accidents and shift control. For a robust false-proof alert system, a precise classification of driving behaviour is needed. However, to the best of our knowledge, the current work lacks complexity, rigidity, synthesised data set, more focus on a particular side of perspective (vehicle, driver, or environment), false-positive classes, and low accuracy.
Drivers can be classified into three groups, namely safe, careless, and dangerous drivers. Capturing driver behaviour is crucial to risk mitigation, and developing context-aware ADAS can influence risk levels and prevent accidents. Moreover, a real-time novel risk assessment determines
Consequently, the key contributions of this study are:
\begin{itemize}
\item	The development of a definition of an adaptive severity level for driver distraction.
\item	A frame-by-frame analysis of driver behaviour severity level in an ADAS.
\item	A proposed model to characterise driver behaviour considering contextual factors such as speed, acceleration, and surrounding vehicles.
\item	Development of the MDDRA model for driving behaviour and its evaluation using machine learning (ML).	
\end{itemize}
The following sections of the paper are structured accordingly.
\begin{itemize}
    \item Section \ref{sec:background} presents the background of risk assessment related to driver's distraction and a detailed literature review on drivers' distraction.
    \item Section \ref{sec:meth} describes the methodology of the MDDRA model, including a rigorous risk assessment.
    \item \ref{sec:exp} presents the experiments and procedures, dataset, and technical setup. 
    \item The findings and discussions of the experimental procedures are presented in Section \ref{sec:results}.
    \item Finally, Section \ref{sec:con} concludes and suggests future work. 
\end{itemize}

\section{Background}\label{sec:background}
We systematically organise the existing literature in relevant areas.

\subsection {Risk assessment for driver'sdistraction}
Risk assessment can be defined as a process evaluating the adverse effects of a natural phenomenon, activity, or substance \cite{b15}. Berdica stated that risk constitutes the likelihood and probability of an incident occurring \cite{b16,b17}. Relative risk ratio is used to quantify vehicle crashing risks under bad weather conditions; its calculation requires a large dataset of crashes arising from adverse weather conditions \cite{b18}. However, using a risk matrix, which combines probability and consequences, overcomes the former method in popularity \cite{b16}. A risk matrix can be used to determine the level of driving risk.  Understandably, most risk indicators related to driver distractions have been modelled after crash events. However, the main flaw in modelling driving risk assessment via post-crash data is that it is a reaction strategy rather than a prevention method. According to Cai et al. \cite{b19}, drivers' subjective assessment of the driving risks - particularly those related to various weather scenarios - is consistent with collision-based studies. In \cite{b19}, the authors assume that the driver's perceived risks are consistent with the actual crash statistics, especially for incidences related to rainy conditions. Various factors can impact driving capability; these can be extracted from the driving context, i.e., from the driver, vehicle, and environment, and include the weather, road, speed, manoeuvres, pedestrians, driver state, and braking. However, there is currently a lack of adequate data and facilities to ensure the development and implementation of an efficient and robust risk assessment model for the driving context. In response to this, this paper proposes using the Naturalistic Driving Study (NDS) TeleFOT, which is sufficiently complete for the environment, vehicle, and driver monitoring. The proposed approach uses the following mathematical model:

\begin{equation}
C_i=\sum_{j=1}^{J}\left(x\cdot X_{i,j}\cdot\beta_j\right)+\varepsilon_i, \quad C_i=1,2\dots,M
\end{equation}

where $C_i$ denotes a discrete model-dependent variable that represents the level of a distraction impact on driving. This variable's various impact levels include minor impact, overall impact, profound impact, and disastrous impact. The $i$ included in this variable represents the $i^{th}$ driver with non-observable $\varepsilon_i$ variables, including the volume of traffic, vehicle type, road type, and rain intensity. A non-observable variable is selected to fit a logistic distribution for generating a continuous latent variable $C_i$ denoting the influence on driving. 

\indent Another proposed approach is the Rank Order Cluster Analysis, which sorts driving risk $R_i$ in ascending order, as indicated by $R_1,R_2,R_3,\dots,R_n$. Consideration of categories $(G)$, including $R_i,R_((i+1) ),\dots,R_j$ and satisfying $j>i$ can be denoted as $G = {i,i+1,\dots,j}$. Consequently, the diameter $D(I,j)$ of $G$ is calculated from the equation:
\begin{equation}
	D(i,j)={\sum_{t=1}^j{(R_t}- R_G) ^2},
\end{equation}
where $R_G$ represents the mean driving risk, and the driving hazard is segmented into $k$ segments expressed as:
\begin{equation}
\begin{array}{c}
G_1= {i_1,i_1,+1,\dots,i_2-1}, \\
G_2= {i_2,i_2,+1,\dots,i_3-1},\\ 
\dots \\
G_k={i_k,k,+1,\dots,i_{k+1}-1} 
\end{array}
\end{equation}
Where the variable $i$ satisfies the following condition:
\begin{equation}
I=i_1<i_2,<\dots<i_k,<i_{k+1}=n+1.
\end{equation}
There is also a minimal loss function with a recursion relationship represented by the equation:
\begin{dmath}
	L[b(n,k)]=\sum_{t=1}^k D(i_t,i_{t+1}-1)	
\end{dmath}
Where $b(n,k)$ denotes a special function returning a classification method based on the values of $n$ and $k$. This function appears in the recurrence relation for the minimal loss function formulated in Equation \ref{eq:lossF} below,
\begin{dmath}\label{eq:lossF}
	L\left[b(n,k)_{k\leq n}\right]=\min_{{k\leq j \leq n}} \left \{L[P(j-1,k-1)]+D(j,n)\right\},
\end{dmath}
where $P(n,k)$ denotes the method to minimise the loss function; if the values of $n$ and $k$ are given; $P(n-1,k-1)$, with different descriptions depicts the optimal driving risk categories. However, our proposed model assumes that driving is a discrete and time-series event; therefore, it takes the risk in the previous frame to compute the severity level of driving risk in the current frame. Furthermore, we consider the sequence of occurrence and the duration of the distracted driving event in computing risk. We address the limitations, thereby enhancing our proposed model.

\subsection{Face Orientation}
Dong et al. \cite{b20} measure driver fatigue based on physiological features, including facial expression and eye activity, as well as the number of times that drivers touch their faces. Assuming that when drivers are tired they exhibit less frequent head motions, it is possible to measure driver fatigue based on the frequency of face turns during a journey; at this moment, the starting duration depends on the deduction from a consecutive number of frames. Hu et al. \cite{b21} state that careless driving is a significant cause of road accidents and tracked driver behaviour using face orientation and facial features taken from infrared images. However, the study uses 
 only a single type of driver distraction, which represents a significant limitation.
 
\indent Sato et al. \cite{b22} infer driver distraction and concentration states using driver body information states. They assessed near misses when drivers approached an intersection. Time series data on different eye-gaze movements and face orientations before the collision or near-miss were logged. Meanwhile, Rasouli et al. \cite{b23} analyzed pedestrian behaviour at crossing points under various weather conditions and road types. Their findings show the vital significance of the head orientation of pedestrians before their crossing intention. Pedestrians make an inference about traffic dynamics (vehicle speed) and crosswalk characteristics (width), and the pedestrian demographics impact their behaviour after the initial purpose of the crossing had been displayed. The result showed an interrelation in the contextual elements, with one factor potentially decreasing or increasing the influence of other factors. 

\indent Fasanmade et al. \cite{b24} used the multi-class distraction to classify driver distraction into severity levels using drivers' physiological features. The approach involved the use of an image-processing rule-based fuzzy logic system. They found that a combination of face orientation and eye glance increases the degree of driver distraction. Furthermore, the results showed that driver distraction could transition from careless driving to dangerous driving when a certain threshold is reached and when multi-class distraction occurs.

\subsection{Hand State}
Hands are vital in driving tasks, such as steering and changing gears. During operation, the state of hands is even more critical in a changing context. Thus, monitoring this state is crucial to the prevention of accidents. Das et al. \cite{b25} describe a naturalistic driving study to detect drivers' hands using bounding boxes and hand annotation. The validation involved checking for false positives that may arise due to illumination conditions, no-hand objects of similar colour, occlusion, and truncation. For the detection, Aggregate Channel Features (ACF) were used as the detector, and the hand detector's accuracy was measured using precision-recall (PR) to evaluate the parameter performance. The initial results suffered from missed detections and false positives, although a cross-dataset comparison yielded better accuracy. 

\indent Dong et al. \cite{b26} state that fatigued drivers assume more comfortable hand positions on the steering wheel. However, Carsten and Brookhius \cite{b73} noted that the impact of cognitive distraction on driving performance differs from that of visual distraction. Specifically, visual distraction adversely impacts a driver's steering ability and lateral vehicle control, particularly car following.
Le et al. \cite{b27,b28} use a novel multiple-scale region-based fully convolutional neural network (MSRFCN) for human region detection in illuminated and low-resolution conditions. They further use a pre-trained network called the ``Oxford'' hand dataset and compare it with several hand detection approaches. The proposed MSFRCN algorithm has an Average Precision (AP) and Average Recall (AR) of 95.1\% and 94.5\%. Besides, it brings an improvement in the AP and AR of 7\% and 13\%, respectively, to classify both the left and right hands.

\subsection{Eye Glances}
According to an eye-tracking study by French carmaker Peugeot, during a one-hour trip in urban traffic, car drivers take their eyes off the road for a total of about two miles. The study involves several drivers on 25 parallel six-mile drives utilising special driver-mounted glasses to investigate where their eyes alighted when operating a range of SUV vehicles. The outcome results found that drivers had eyes off the road for about 7\% of the trip \cite{b29}.

\indent Yuan et al. \cite{b30} suggest an approach to classify existing driving situations and forecast off-road vehicle situations using a Hidden Markov Model (HMM). The experiment is carried out using a driving simulator analysis involving 26 participants in three driving scenarios: rural, urban, and motorway environments. Three different occlusion durations ($0 s$, $1 s$, and $2 s$) are added to measure the eyes-off-road period durations. The results reveal that existing driving situations can be optimally defined using glance position sequences, with up to 89.3\% accuracy. The motorway is the most distinguishable, with over 90\% precision. Moreover, in driver's eyes-off-road period estimation, using HMM-based algorithms with two inputs (namely look duration and look position sequence)s, gives the highest accuracy rate of 92.7\%. Vehicles of 42 newly approved adolescent drivers are fitted with sensors, accelerometers, and Global Positioning Systems (GPS) to collect data continuously for 18 months period. Crash and near-crash (CNCs) situations were reported through the investigation of significantly elevated gravitational force incidents. Analysis of the video assigns 6 seconds previous to each CNC, and random samples of non-CNC road fragments are coded for the period of eye-glances off the front road and occurrence of secondary mission participation. The likelihood (odds ratio) of CNCs due to eye-glance activity is determined by contrasting the prevalence of secondary task participation and the length of off-road eyes before CNC with the prevalence and period of off-road looks non-CNC road segments. The crash incidence improved with the period of the single-most prolonged glimpse during all secondary tasks (OR$ = 3.8$ for $>2 s$) and wireless secondary task presence (OR $= 5.5$ for $>2 s$). The single-most extended glimpse offers a constant estimation of an accident's likelihood than absolute eyes off the forward roadway \cite{b31}.  

\subsection{Road Type (Urban, Highway)}

In \cite{b32}, Doshi et al. develop an algorithm that includes critical vehicle data, such as the brake switch status, throttle position, and wheel speed. It also uses the inputs to calculate several parameters, namely shifts per given time interval, throttle variations, mean velocity, and acceleration. The resulting parameters help the algorithm to identify the road type on which the assigned vehicle was travelling. This road identification process is achieved through parameter comparisons with reference values that defined various road types. Additionally, the algorithm identified the driver type using driver inputs, such as gear shift patterns and the driver's handling of the brake and accelerator pedals. Doshi et al.'s algorithm attained a Receiver Operating Characteristic (ROC) value of 85\% accuracy in road-type identification.

\indent The behavioural analysis on road rage in China by Chai et al. \cite{b33} reports an inverse proportionality between cases of road rage and lane number on a given road. In other words, with fewer lanes, there are more incidences of road rage. Additionally, the study reveals that road rage increases with an increase in the number of nonmotorised vehicles. Road rage generally involves fewer trucks, although more trucks are involved in road rage on highways and daytime driving, which leads to fewer incidences involving non-motorised vehicles. A limitation of the study is its small sample size and lack of demographic and environmental variables, which require a more detailed analysis in the future. To characterise road types and measure the degree of driver aggression, Messeguer et al.

\indent \cite{b34} design and implement a neural network-based algorithm to assist drivers by highlighting unacceptable driving behaviours. Their test results demonstrate the ability of neural networks to achieve a degree of precision in the classification of driver and road types. 
Since contextual information plays a critical role in the accurate performance of various road classification and driver distraction identification algorithms, useful contextual information collection is vital.

\indent Rakotonirainy et al. \cite{b35} and Khan \cite{b36} propose a context-aware system for the real-time collection and analysis of contextual information related to a vehicle, the immediate environment, and the driver. The system also gathered information from questionnaires. A Bayesian network model is employed to analyse contextual information through a learning model. This facilitates the observation and prediction of a driver's future moves. The model attains a comparable accuracy in predicting future driver behaviours and warning other road users. However, the system is too complicated to implement in real life; besides, false alerts are a key problem of this system.

\indent Methods for recognising and classifying road traffic accident severity play an essential role in understanding accidents, their causes, and possible mitigation strategies. To that effect, Jianfeng et al. \cite{b37} design a set theory-based accident recognition and classification method that supports vector machines. Their model employs rough set theory in calculating the significance of the driving environment, road, vehicle, and human attributes, and the results demonstrated the model's ability to improve recognition accuracy and reduce computational workloads. The study's limitation is that human physical behaviour, such as the driver's face orientation, is not considered.

\subsection{Weather}
 The travel weather warning system (TWWS) by Cai et al. \cite{b19} is similar to a Road Weather Information System (RWIS) for sharing weather safety information and disseminating safety warnings to drivers. This system is made up of risk estimate models that were based on extensive weather-related crash data. Weather-related data are collected using questionnaires, wherein drivers identify the various risks encountered while driving under different driving conditions. The severity of each type of weather is measured on a four-point scale ranging from slight to catastrophic. Metrics such as the intensity of rain and traffic volume are also considered.
 
 \indent Malin et al. \cite{b38} state that rainy weather is a significant factor in traffic accidents, with the risk of accidents increasing under poor road weather conditions.  
 
\indent Sherretz and Farhar \cite{b39} report a positive linear correlation between rainfall and the frequency of road traffic crashes. Bergel-Hayat et al. \cite{b40} also revealed a significant correlation between the aggregate number of traffic accident injuries and weather variables. They observe that the correlation between these two parameters varies depending on road type.

\indent Brodsky and Hakkert \cite{b41} propose measuring the risk of a road accident during rainy weather. Their method shows a drastic rise in road traffic accident injuries during rainy weather compared to dry weather. The increased dangers under wet conditions that follow a long dry season are well known to drivers, as was found by Knapper \cite{b42}. 

\subsubsection{Speed}
Maintaining the correct speed continues to be a challenge for many drivers. Drivers who violate driving rules, such as speeding, are said to crash more often. Stradling and Auberlet \cite{b28}, \cite{b43} show that vehicle trajectory variations may reveal valuable details on how spatial restrictions impact driver behaviour (e.g., lateral location and speed). On the one hand, the findings reveal that while driving on the vertical curve of the crest, before encountering oncoming vehicles and narrower lane width, the lateral location variability is more significant. However, this was reduced according to the perceptual procedure used. Another study investigated the impact of multiple factors, such as image size, speed, road shape, driving experience, and gender, on the perception of speed by drivers.

\indent Wu et al. examine the most miniature image scale (38\% of the actual field of view) in \cite{b44} and find that speed calculations are the most reliable. Driving velocity is gradually undervalued as image scale grows. Participants with driving experience correctly measured driving speed on both wide and narrow roads. However, those without driving experience make more underestimates on broader roads. Furthermore, the environmental conditions concerning speed performance have been highlighted by Bellis et al. \cite{b45}. They challenge current policies and suggest interventions by teaching drivers about the inverse illumination-speeding relationship and measuring how better vehicle headlights and intelligent road lighting can attenuate speed. 

\indent Real-world speeding and its association with illumination, an environmental property described as the incidence of luminous flux on a surface, have been examined in the literature. Manser and Hancock \cite{b46} address the need to determine whether visual patterns and wall tunnel texture impact driving performance since maintaining the correct speed continues to be challenging. The findings show that the relationship between driver speed and reaction is affected by the visual pattern and the texture of the tunnel wall.

\subsection{Vehicle}
Mishra and Baja \cite{b47} and Kamar and Patra\cite{b48} adopt ML to predict the driving patterns of drivers and the impact on their social behaviour using CCTV cameras installed to monitor traffic. Their observations are carried out during the day and the metrics used for measurement are instances of traffic violations due to aggressive patterns.

\indent Lee and Kum \cite{b49} propose a feature-based lateral position estimation algorithm that employs lateral positioning and stereo vision, regardless of changes in viewpoints and obstructions, resulting from pixel-wise feature extraction. The algorithm extracts vehicle images through image filtering and thresholding, and removes the ground portions from images captured by cameras. The algorithm's detection component consists of a deep convolutional neural network with a speeded-up robust feature (SURF) to match successive image frames. They estimate the lateral position of ground points using an inverse perspective mapping algorithm (IPM). The testing and validation phases are performed using urban and highway methods to achieve zero mean error and a standard deviation of $0.25 m$ in the estimation of the lateral position. 

\indent \cite{b50} detected driver behaviour based on car follower behaviour, which can vary according to distraction, fatigue, driver habits, and surrounding traffic. On-road trajectory data obtained in Beijing are used in their study, and distinctive driver states and car-following models are observed as metrics. This led to the prediction of the driver's velocity control with improved accuracy. 

\indent Mittal \cite{b1} used object detection and a faster R-CNN model to detect vehicles of different scales and sizes. An evaluation is performed using the $FLIR_{ADAS}$
dataset for both RGB and thermal images. Gong et al. \cite{b52}instead propose using the YOLOv3 algorithm to detect vehicles in thermal images, achieving a 65\% higher accuracy and speed than the original YOLOv3-tiny.

\subsubsection {Pedestrians}
Kharjul et al. introduce in \cite{b53} an active protection automobile pedestrian identification device to minimise the amount and intensity of vehicle-pedestrian collisions. They present a pedestrian identification approach that segments pedestrian candidates in images. The method uses Ada-Boost and cascading algorithms to confirm whether each claimant is a pedestrian. The Support Vector Machine (SVM) is used as a final classifier which exploits the input features  of grey images for training. 

\indent Taiwan and Yamada develop in \cite{b54, b55} a tool for calculating driver knowledge and behaviour concerning pedestrian positions at crosswalks and while crossing, especially at left and right turns at intersections. Their findings on an appraisal carried out using objective evidence of driving behaviour on public roads have also been published.  In contrast, Rangesh et al. \cite{b56} examine the behaviour of pedestrians. In particular, from a solely vision-based point of view, they concentrate on detecting pedestrians engaging in secondary behaviours involving their mobile phones and other handheld multimedia devices. They propose a pipeline integrating articulated human pose prediction and gradient-based picture features to detect the presence/absence in either a pedestrian's hand. A belief network is used to encode knowledge from multiple streams and their dependency on each other. This network is then used to forecast a likelihood score that suggests pedestrians' engagement with their devices. 

\indent Phan et al. \cite{b57} focus on drivers' actions whenever a person emerges in front of their car. They used two static parameters-based methods, namely the Necessary Deceleration Parameter and Time-To-Collision, and compare them to the proposed approach. They also employ a Hidden Markov Model (HMM) to characterise driver knowledge and unawareness of pedestrians. Compared with the baseline algorithms, the outcome indicates a significant enhancement of the HMM-based process.

\subsection{Illumination (Day, Night)}

In \cite{b58}, Clarke et al. observe that the rate and severity of road traffic accidents are influenced by driving. In their study, the visibility conditions under investigation include rainy and night driving, with the control test being dry daytime driving. Their findings on the increased rate and severity of crashes at night and during rainy weather match the conclusions of \cite{b59}, where the risk of fatal crashes is shown to increase by a factor of four during night driving compared to daytime driving.

\section{The multi-class driver distraction risk assessment model} \label{sec:meth}

Based on existing literature, we develop and test our hypothesis that driver behaviour based on driver distraction has different severity levels, which we define as `safe', `careless', or `dangerous'.

We then justify the weighting metrics for the distractions present in the TeleFOT dataset \cite{b67}. The following observable parameters can characterise signs of attention deficit and fatigue in the driver: PERCLOS (PERcentage of eye CLOSure, i.e., the percentage of the time the driver's eyes are closed) \cite{b60}, turning the head to the left/right to the body, tilting the head forward relative to the body (the moment when the driver is ``nodding off''), duration and frequency of blinking, and the degree of openness of the person's mouth (a sign of yawning). In particular, for PERCLOS, there was a discrete number of parameters defined, namely P70, which is the proportion of time for which the eyes were closed of at least 70\%; P80, which is the proportion of time for which the eyes were closed of at least 80\%; and EYEMEAS (EM), which is the mean square percentage of the eyelid closure rating \cite{b60}. Furthermore, general information describing a driver helps not only explicitly identify that driver among all other drivers who installed and used a particular monitoring software package but also helps to improve the search for and classification of drivers with similar characteristics (general patterns among groups would help to predict developing situations). This can be accessed via the database, with a weight coefficient applied since this is a ``common'' behaviour rather than an individual driver's behaviour. Ginting et al. \cite{b61} adopted a 5-point Likert scale to model anxiety about individual coronary heart disease at different levels. Lopez-Fernandez et al. \cite{b62} also used a scale in assessing problematic internet entertainment among adolescents. The scale adopted was a self-administered scale for measuring the behavioural addiction of online social network users and video gamers regarding the degree of severity. Based on this, we formulated the distraction severity levels. At this moment, the ratings of the severity level of distractions were designed using a 5-point Likert-type scale, as seen in Table \ref{tab1}, \cite{b63, b64,b65}.

The proposed model considers the severity level of driver distraction based on an observation of their driving history. While this can be unpredictable, we analyse the driver's behaviour frame by frame to obtain intricate details. We take the following steps:
\begin{itemize}
    \item Decompose the video to a frame-by-frame level.
\item Study each frame to assess its severity level.
\item Aggregate the previous severity level of frames to the current frame severity level.
\item Provide a precise class of severity based on the calculated severity level.
\end{itemize}
In the following, we outline the essential aspects of our model for accessing the severity level of driver distraction. We acquired the knowledge and data by observing and analyzing individual frames from the input source. We began by formulating the risk assessment based on driver behaviour according to $P = {p_1, p_2, p_3, \dots,p_n}$, as described in Table \ref{tab1}. Each parameter $P_i$ is characterized by some set of action $A = {a_1,a_2,a_3,\dot,a_n}$, with each action $a_i$ having a weight $W_i$.

\begin{table}[ht!]
	\caption{The essential aspects of the proposed model for accessing the severity level of driver distraction. The maximum weight values are reported in boldface.}
	\label{tab1} \centering
 
	\begin{tabular}{|c|c|c|}
		\hline
		\textbf{Parameter}                     & \textbf{Action}                 & $\mathbf{W_i}$ \\
		\hline
		\multirow{3}{*}{State of Hand}    & Double hand      & 0      \\
		&                    Single hand            & 1      \\
		&                     No hands               & \textbf{2}      \\
  \hline
		\multirow{3}{*}{Road Type}        & Urban                  & 1      \\
		&                     Dual                   & 2      \\
		&     Highway           & \textbf{3}      \\
    \hline
		\multirow{2}{*}{Face Orientation}  & On road         & 1      \\
		&                     Off-road               & \textbf{2}      \\
    \hline
		\multirow{2}{*}{Illumination}     & Day                    & 1      \\
		&                     Night                  & \textbf{2}     \\
    \hline
		\multirow{3}{*}{Eye Gaze}        & Eyes on road           & 0      \\
		&                     Eyes off-road          & 1      \\
		&                     Eyes shut              & \textbf{2}      \\
    \hline
		\multirow{3}{*}{Weather}          & Dry                    & 1      \\
		&                     Rain                   & 2      \\
		&                     Snow                   & \textbf{3}      \\
    \hline
		\multirow{3}{*}{Manoeuvres}       & Stopped                & 0      \\
		&                     Turning                & 1      \\
		&                     Moving                 & \textbf{2}      \\
    \hline
		\multirow{2}{*}{Surroundings}      & Vehicle not present    & 0      \\
		&                    Vehicle present        & \textbf{1}      \\
    \hline
		\multirow{2}{*}{Pedestrians}                                      &  not present & 0    \\ 
    &  present & \textbf{1}    \\ \hline
	\end{tabular}
\end{table}
The next stage consists in identifying the severity levels, according to severity rates, respective colour for identification, and classification label to start with. For instance, if the severity is 0.0, the risk colour will be right green, this will be no distraction from the driver has been observed, and it will have no impact on the driver's life. While, if the severity level is 0.9 or above, the risk colour will be red, and it will mean that a severe causality can be expected, and it is hazardous to keep driving. Table \ref{tab2} provides these details along with the relevant consequences.

\begin{table}[ht]
		\caption{ Severity level, the risk colour and its impact on the consequences}
	\label{tab2}
	\centering
		\begin{tabular}{|c|c|c|c|c|}
		\hline%
		\textbf{Severity}  & \multirow{2}{*}{\textbf{Risk Color}}  &\multirow{2}{*}{ \textbf{Impact}} & \multirow{2}{*}{\textbf{Distraction} }    & \multirow{2}{*}{\textbf{Consequences} }                                                         \\ 
  (0.0 - 1.0)&&&&\\
  \hline
		0.0                                                                  & Light Green & No Impact       & Safe                                                                & No distraction                                               \\ \hline
		0.1-0.25                                                             & Green       & Slight Impact   & Safe                                                                & Slight distraction                                           \\ \hline
		\multirow{2}{*}{0.25-0.399}                                                           & \multirow{2}{*}{Yellow}      & \multirow{2}{*}{Low}             & \multirow{2}{*}{Safe}                                                                & Noticeable    \\
 &       &              &                                                                 &  distraction    \\ \hline
		0.4-0.599                                                            & Dark Yellow & Medium          & Careless                                                            & distraction detected                                         \\ \hline
		0.6-0.79                                                             & Orange      & High            & Dangerous                                                           & Frequent distractions                                         \\ \hline
		0.8-0.9                                                              & Dark Orange & Very High       & Dangerous                                                           & Casualty prone                                                        \\ \hline
		\multirow{2}{*}{0.9-1.0}                                                              & \multirow{2}{*}{Red}         & \multirow{2}{*}{Extreme}         & Extremely  & Severe  \\ 
     &          &          & Dangerous &  casualty Prone\\ 
  \hline
	\end{tabular}
\end{table}

\subsubsection{Risk Assessment Matrix}
An approach to the computation of risk assessment in a quantitative model uses a Risk Assessment (RA) Matrix's graphical tool. The risk matrix involves calculating the magnitude of the potential consequences scaled on the vertical axis (levels of probability) of these consequences occurring; technically, the probability of these consequences occurring on the horizontal axis. This facilitates an increase in the visibility of risk and impacts decision-making.  The risk is computed by calculating the Consequence Likelihood of Occurrence Likelihood. Here, the likelihood depicts the probability of a driver's distraction being related to their context awareness. Consequences/Severity Level: The occurrence of multi-class context-aware distractions is classified into severity levels of distraction. 
\subsubsection{Probability}
Probability is the measure of the likelihood that an event will occur. For example, there is a possible aggregation to measure the number of times a driver experiences a particular distraction during a driving course. The driver may be profiled according to the distraction severity level at the end of the driving course. 
\subsubsection{Likelihood}
The likelihood levels can be described as frequency values (duration course) and state values (every frame). Four impact levels are considered in this paper, namely no impact, low impact, medium impact, and high impact. When an effect has no impact, the likelihood score is one, and the likelihood of that distraction observes no distraction or a distraction that has not currently occurred. When a slight distraction is detected, the impact is low, with a score of 2. A medium result is considered when a minor distraction has occurred, and the score is then set to 3; 4 implies a medium to significant distraction occurrence. More impacts can be seen in Table \ref{tab3} below.

\begin{table}[ht!]
		\caption{Severity Risk Matrix, 5 continuous frames, progression of danger}
	\label{tab3}
 \centering

 \begin{tabular}{|c|c|c|c|c|c|}
 \hline
 \textbf{Severity} & \multicolumn{5}{c|}{\textbf{Severity values}}\\
  \hline
 
		\hline
		Extreme         & 7 & 7 & 14 & 21 & 28 \\ \hline
		Very High       & 6 & 6 & 12 & 18 & 24 \\ \hline
		High            & 5 & 5 & 10 & 15 & 20 \\ \hline
		Medium          & 4 & 4 & 8  & 12 & 16 \\ \hline
		Low             & 3 & 3 & 6  & 9  & 12 \\ \hline
		Slight/Very low & 2 & 2 & 4  & 6  & 8  \\ \hline
		No Impact       & 1 & 1 & 2  & 3  & 4  \\ \hline
	\end{tabular}
\end{table}

We propose using a weighted average of the parameters to compute the severity levels per frame, as depicted in Table \ref{tab1}-\ref{tab4}. These weights are capped by the maximum number that a parameter can take. For example, we take ``State of Hand'' as a parameter and grade it as follows: 0 - double hands, 1- single hand, 2- no hands. If the value of a given frame for this parameter is x, then the weighted value is x/2 since the maximum value, of this parameter, can take 2.  Let us generalise this for any parameter $x_i$ with a maximum value $m_i$  as 
$\text{Severity level}= \sum_{i=0}^{n}x_i/m_i$. Where $n$ is the number of parameters in consideration.
\begin{table}[ht!]
	\caption{Severity Risk Assessment Matrix}
	\label{tab4}
 \centering
	\begin{tabular}{|c|c|}
		\hline
               	 \textbf{RA} & \textbf{Likelihood}                                 \\ \hline         
		1                        & No distraction is observed or occurred yet \\ \hline
		2                        & A slight distraction has been observed     \\ \hline
		3                        & A minor distraction has occurred           \\ \hline
		4                        & A medium or major distraction has occurred \\ \hline
	\end{tabular}
\end{table}
\subsubsection{Special considerations}

To ensure the safety of drivers \cite{b66} developed a lane departure warning system based on image processing using a mono camera installed inside the car. A distinctive feature of the system is that it successfully processes several road conditions, including undesirable situations such as changing the width of the road lane, the radius of its curve, the direction of the road, and the complete absence of a road surface. From this we gather that speed depends on the road type; hence, we multiply it by the weight of its road for speed. We considered road type in the UK as this conforms to the source of the dataset. For the metric of road types, we define the threshold according to the speed limit allowable on the road type, i.e., urban, single carriage, and motorway at 30 mph, 60 mph, and 70 mph, respectively.  Furthermore, we define the following contextual data:
\begin{itemize}
\item Vehicle $V$ and driver data with probabilities $P(V)= {v1,v2,\dots,v_m}$
\item Environment $E$ and environmental data with probabilities $P(E)= {e1,e2,\dots,e_n}$
\item Speed $a$
\item Probability of Surrounding $P(S)$ 
\item Probability of Pedestrians $P(Pe)$ 
\end{itemize}

We formulate the following equations. The speed is computed as described in equation \ref{eq:speed}, for example, given that the national speed limit of the UK is 70 mph and the maximum road type weight and the score is 3:

Let the national speed limit of the UK be denoted by $MaxSpeed = 70$ mph, and let the maximum road type weight and score be denoted by $MaxRoadType = 3$. Then the speed on a given road can be computed as follows:

\begin{equation}\label{eq:speed}
Speed = \frac{MaxSpeed \cdot MaxRoadType}{RoadType}\cdot\frac{1}{\frac{RoadType}{MaxRoadType}}
\end{equation}

where $RoadType$ is the numerical value representing the type of road, such that higher values indicate more challenging road conditions. The fraction $\frac{RoadType}{MaxRoadType}$ adjusts the speed based on the road type so that the speed is reduced on more challenging roads.

We understand there are different data points in each frame; thus, the severity level of a given frame with k data points where severity $(S^*)$ of a given frame is $(fi)$:	
\begin{equation}				
S(f_i)=  \frac{1}{k}\left(\sum_{i=0}^{m} P(V_i) +\sum_{j=0}^{n} P(E_i)+a P(S) + P(Pe)\right)	
\end{equation}

We now compute the aggregate severity $(S^*)$ of a given frame $(fi)$ given the last $i-t$ frames. This is achieved by taking the average of the current frame's severity score compared to the severity score of the last i-t frames:
\begin{equation}
S(fi)=\frac{1}{t}\left((S(fi)+ \sum_{j=t}^{i-1} S(f_j) \right)
\end{equation}

The verification and validation processes for the proposed model typically include both computational and physical aspects. To assess the degree of adequacy of the numerical modelling, the following steps can be performed: 1) Determine the order of convergence of numerical solutions in comparison with a numerical solution using a reduced number of parameters; and 2) assess the sensitivity of the sampling algorithm to various uncertainties, including parameter constraints, grid adaptation to real measurements and boundary conditions. Furthermore, validation assumes a careful comparison of the numerical calculation results of the phenomenon under study with experimental data to obtain an answer to the question ``is the numerical solution, correct?''. Thus, a comparative analysis of the model with all the conditions, including the uncertainties associated with missing parameters and boundary conditions from the real world and computational points of view, is carried out.

A few methods can be used for model validation purposes; 
\begin{itemize}
    \item The L2 Loss method consists in evaluating the loss function to compute the squared error for each training dataset, thus returning the square of the differences between the actual and the predicted values, i.e. $ L=\left(y-f(x)\right)^2$.
\item Errors (residuals) can be predicated via a cross-correlation test: \textit{Are the residuals uncorrelated with the input?}

\item  The model can be applied to unseen data (cross-validation). This strategy may be helpful since it establishes the robustness of the proposed model. It can also provide the basis for the hybrid cross-correlation validation since there is a need to separately investigate how the inputs and outputs are correlated and how this correlation is affected by our modelling scheme;

\item ``Inverse Problem'' approach, i.e., acquire a solution to the problem and solve the inverse case to obtain the output parameters. This will help to validate the assigned weight coefficients and the overall parameterisation scheme. 
\end{itemize}

In our case, the reliability of our model is tested by performing a cross-correlation test to analyze the residuals. This is carried out on the data obtained over two separate datasets, with the analysis, separately applied to the inputs and outputs.

\section{Experimental methods} \label{sec:exp}
\begin{figure}[ht!]
	\centering
	\includegraphics[width=0.8\linewidth]{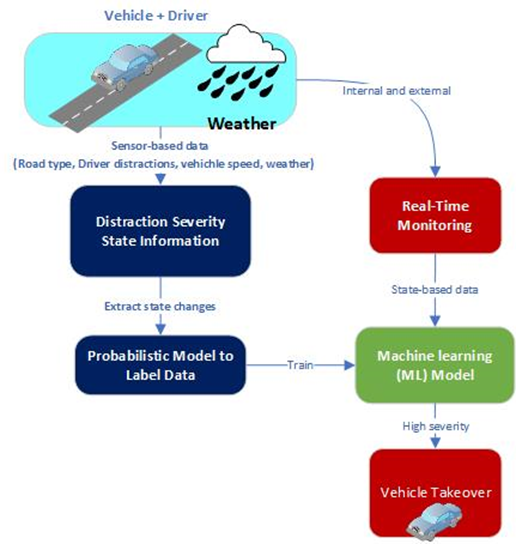}
	\caption{MDDRA Model Architecture}
	\label{fig:fig1}
\end{figure}

A discrete-time model is proposed for the application of ML to detect the pattern in time-series driver distraction data. Consequently, we develop an adaptive model for the prediction of the driver's severity level based on distraction. The MDDRA model architecture illustrated in \ref{fig:fig1} the state flow of the data and system modules that constitute the entire system. The architecture is made up of six states, as described in the following subsections.

\subsection{Driver and vehicle data collection}
All required data are collected (from the vehicle and driver) using multiple sensors and video recorders. Sensor-based data are collected in real-time, including road type, driver movements, face and head direction, vehicle speed, weather, and the surrounding driving environment. 

\subsection{Object extraction}
This architecture module extracts distraction state information (gaze at something else, Overspeed, etc.), including the changing state of the distraction, and feeds it into a probabilistic model for labelling. 

\subsection{Data labelling}
The probabilistic model is further applied to the labelled extracted data, which is then used to train the system's core engine before the ML model is applied.

\subsection{Real-time monitoring}
Context-aware real-time data from the real-time driving video streams of the internal and external sensors of the vehicle are monitored. The data are further analyzed, and feature extraction of both the driver and vehicle state-based data is performed; this is then fed into the ML model.

\subsection{ML model}
The ML model takes in state-based data (eye gaze, state of hands, speed, face orientation, manoeuvre) and training datasets to predict the level of distraction. The resultant model is the probability of the occurrence of driver distraction in the current distraction frame state $P(C_{t+1} S_{t+1})$,  measured as the state transition from the previous frame state, denoted as $P(S_{t+1} C_t)$. If the severity of distraction is high, vehicle takeover operations take effect.

\subsection{Vehicle takeover}
The severity informs the decision to perform a vehicle takeover of the distraction detected by the ML. If the distraction passes the threshold, i.e., transitions from careless to dangerous, then the decision for the vehicle to take over driving is triggered.

\begin{figure}[ht!]
	\centering
	\includegraphics[width=\linewidth]{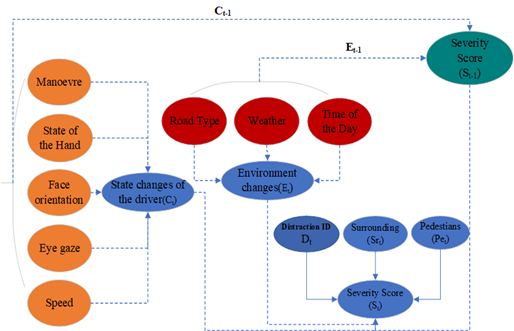}
	\caption{Context-aware probabilistic model for severity classification}
	\label{fig:fig2}
\end{figure}

A dynamic Bayesian network (DBN), as depicted in Figure \ref{fig:fig2}, is an extension of a Bayesian network that uses the time (dynamic) concept in modelling sequential time-series observations. It also uses a probabilistic inference model in handling uncertain information. An acyclic graphic represents the conditional independent and latent temporal variables discretely and continuously. For this case, the inference from the DBN is derived from three key classes of nodes. Namely, driver features nodes, distraction identifiers, and contextual data. These inputs are represented in this model by nodes such as the state changes of the driver, consisting of 5 central nodes, namely face orientation, speed, manoeuvres, eye gaze, and state of hands. The environmental changes node, consisting of road type, weather, and time of the day, forms part of the contextual input data into the model and data on pedestrians and the surrounding environment. The final input is the distraction identifier derived from the analysis of the driver features by a hybrid CNN-LSTM. The output of this acyclic graph is a severity score, which is the measure of the degree of the distraction of the driver extracted from the driver features and contextual information.

\subsection{Dataset}
\begin{figure}[ht!]
	\centering
	\includegraphics[width=\linewidth]{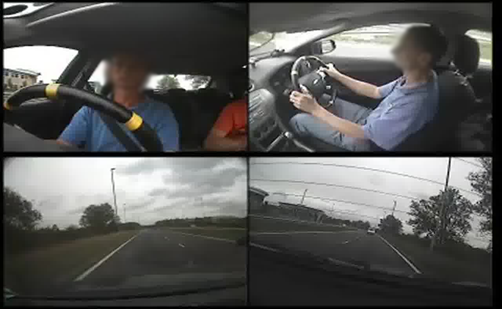}
	\caption{TeleFOT Dataset}
	\label{fig:fig3}
\end{figure}

The TeleFOT Naturalistic driving study dataset is a European Field Operation Test (FOT) \cite{b67}. The project was designed to enhance research on intelligent transportation systems \cite{b68, b69}. The experiment was conducted in the UK and involved 27 participants \cite{b69}. Each driving video consists of four video channels that monitor in-vehicle and out-vehicle parameters, including face orientation, eye gaze, and hand position. The dataset consists of time-series data, the camera is located inside the vehicle on the dashboard and on the passenger side of the vehicle, it also has two extra cameras to capture the environment as can be seen below illustrated in \ref{fig:fig3}.

\subsection{Probabilistic data model}
We considered the driver distraction state's changes frame by frame, as depicted in Figure \ref{fig:fig2}. Technically, our proposed Context-aware probabilistic model for severity classification can be described as the probability of the occurrence of driver distraction in the current frame state $P(C_{t+1} \| S_{t+1})$ from the previous frame state $P(S_{t+1} C_t )$. So, there exists the probability of the occurrence of distractions in the environmental state $P(E_{t+1} \| S_{t+1})$, if the current frame state $S_(t+1)$ changes from the previous frame state $S_t$. The proposed extended dynamic Bayesian model includes several environmental variables such as road type, weather, and day. In order to compute the probability severity scores $P(S_{t+1}\|C_{t+1},E_{t+1},Sr_{t+1},Pe_{t+1},S_t )$, we utilized the dynamic linear model \ref{eq:11} and which is a combination of the state change of driver distraction $C_{t+1}$, environmental changes $E_{t+1}$, distraction identification $D_{t+1}$, pedestrians $Pe_{t+1}$, and surroundings $Sr_{t+1}$.
\begin{equation}\label{eq:11}
\begin{split}
P (S_{t+1}\|C_{t+1},E_{t+1},Sr_{t+1},Pe_{t+1},S_t )=  \\
P (C_{t+1}\|S_{t+1} ).P(E_{t+1}\|S_{t+1} ).\\P(Sr_{t+1}\|S_t ).P(Pe_{t+1}\|S_{t+1} ). \\P(D_{t+1}\|S_{t+1} ).P(S_{t+1}\|S_t)
\end{split}
\end{equation}

\subsection{Interdependencies Test}

\begin{figure}[ht!]
	\centering
	\includegraphics[width=\linewidth]{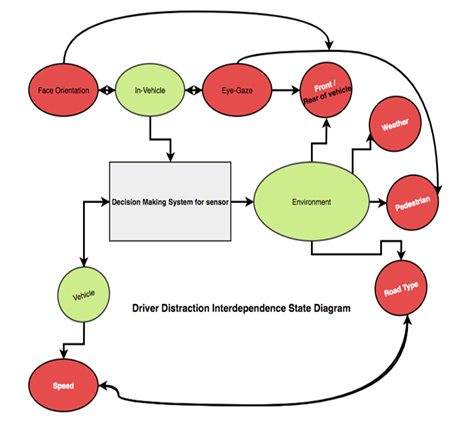}
	\caption{Distraction Interdependence State Diagram}
	\label{fig:fig4}
\end{figure}

We can apply the developed interdependency test for road type and its impact on driving speed. For example, in Figure \ref{fig:fig4}, the regression analysis coefficient is calculated as 0.529134, implying a significantly positive relationship. We assume that the driver would drive within the UK speed limit. The dataset of the driver may be more biased towards a degree of severity compared to other databases. Thus, it is necessary to validate the model using a regression model to test the interdependencies. We perform a correlation analysis between driver distraction and the severity classification of the distraction. Also, we conduct a multi-linear regression analysis to estimate the influence of driver distraction on the degree of severity classification.

\subsection{Data Normalisation}
We log and normalize the vehicle's speed synchronously with the distraction state frame; thus, the time-series data of the vehicle are correlated at every frame. Subsequently, a regression analysis is conducted to validate our hypothesis, as seen in the results section. The severity level classification of the baseline drivers is likely to have a lower mean than experienced drivers. Furthermore, a regression analysis is performed to indicate the mean of the in-vehicle parameters, mean vehicle data, and mean environmental data to produce the safe severity level. Meanwhile, in professional drivers, a safe severity level is likely to be more than the baseline. However, in other parameters, like per frame, severity means an aggregate severity means. In contrast, the statistical analysis of all the parameters contributes significantly toward the severity level considered safe, careless, or dangerous. However, only the vehicle speed distribution across the journey and its relation to the road type and driver distraction severity level has a substantial impact.

\subsection{Results of the model validation procedure}

To provide basic information about the variables in the dataset, the descriptive statistics for one of the simulated events (driver 1, event 1) are presented in table \ref{tab5}. The mean, median, kurtosis, and skewness values are calculated by using equation ~\eqref{eq:12} to~\eqref{eq:15}.

\begin{equation}\label{eq:12}
\text{mean}= \frac{Sum (x)}{Count (n)}
\end{equation}
\begin{equation}\label{eq:13}
\text{Median}(x)=\begin{cases}
	X\left [ \frac{n}{2} \right ] & \text{ if } n= even\\
	\frac{X\left [ \frac{n-1}{2} \right ]+X\left [ \frac{n2+1}{2} \right ]}{2}& \text{ if } n= odd
\end{cases}
\end{equation}
	Where X is an ordered list of values in the data set, and n denotes the number of values in the data set (count). For a univariate data $y_1,y_2,\dots,y_N$ the formula for skewness is:
	\begin{equation}\label{eq:14}
g_1= \frac{\sum_{i=1}^{N}\frac{\left ( y_i-\overline{y} \right )}{N}^3}{S^3}
\end{equation}
	Where $ \overline{y}$ shows the mean, $S$ is the standard deviation, and $N$ is the number of data points. Note that in computing the skewness, the s is computed with $N$ in the denominator rather than $N-1$. 
	
\begin{equation}\label{eq:15}
kurtosis = \frac{\sum_{i=1}^{N}\frac{\left ( y_i-\overline{y} \right )}{N}^4}{S^4}
\end{equation}

	\begin{equation}\label{eq:16}
S= \sqrt{\sum \frac{\left ( y_i-\overline{y} \right )^2}{N}}
\end{equation}
	
The results of these metrics suggest that this is a symmetrical distribution. This reflects how the data were modelled. It would be valuable to deploy this model using real data from the video sensor to access the accurate distribution of parameters, such as face orientation and eye gaze, and then analyse the results.
\begin{table}[ht!]
\centering
	\caption{Descriptive Statistics}
	\label{tab5}
	\begin{tabular}{|l|l|}
		\hline
		Mean                      & 0.513049625 \\ \hline
		Standard Error            & 0.007304311 \\ \hline
		Median                    & 0.508023896 \\ \hline
		Standard Deviation        & 0.118456024 \\ \hline
		Sample Variance           & 0.01403183  \\ \hline
		Kurtosis                  & 0.057262351 \\ \hline
		Skewness                  & 0.217203269 \\ \hline
		Range                     & 0.725482175 \\ \hline
		Minimum                   & 0.162361126 \\ \hline
		Maximum                   & 0.887843301 \\ \hline
		Sum                       & 134.9320515 \\ \hline
		Count                     & 263         \\ \hline
		Largest (1)               & 0.887843301 \\ \hline
		Smallest (1)              & 0.162361126 \\ \hline
		Confidence Level (95.0\%) & 0.014382625 \\ \hline
	\end{tabular}
\end{table}

To validate the model, its predictions are tested using correlation analysis, as suggested in Section \ref{sec:exp} This technique is typically used to test relationships between quantitative or categorical variables. Correlation coefficients have a value between -1 and 1. A ``0'' value means that there is no relationship between the variables, while -1 or 1 means a perfect negative or positive correlation (negative or positive correlation here refers to the type of graph the relationship will produce).

\begin{table}[ht!]
\centering
	\caption{Correlation Coefficients}
	\label{tab6}
	\begin{tabular}{|l|l|}
		\hline
		State of Hand    & 0.425847 \\ \hline
		Road Type        & 0.363796 \\ \hline
		Face Orientation & 0.420461 \\ \hline
		Time of day      & 0.224532 \\ \hline
		Eye Gaze         & 0.296584 \\ \hline
		Weather          & 0.247372 \\ \hline
		Maneuver         & 0.323121 \\ \hline
		Speed            & 0.053056 \\ \hline
		Surrounding      & 0.441935 \\ \hline
		Pedestrians      & 0.255076 \\ \hline
	\end{tabular}
\end{table}

From Table \ref{tab6}, it is clear that there is a positive correlation with all but one of the parameters used in the model, that is, the speed of the vehicle.

The model is also tested on multiple events, and the results demonstrate a consistent lack of correlation with velocity. This might indicate a need for a wider velocity span to be present in the dataset or, if this does not affect the results, better represent the model's influence.
\section {RESULTS AND ANALYSIS}\label{sec:results}
The implementation of our model and architecture in Figure \ref{fig:fig1} was carried out to determine which ML model will best predict driver distraction to aid in vehicle takeover decision-making. Furthermore, to avoid bias, the results of the experiment were determined using different ML algorithms in the dataset. We analyse the scatter plot and the confusion matrix of the predicted class. 
\subsection {INTERDEPENDENCY TEST USING REGRESSION ANALYSIS}
The regression analysis was performed on the following three context-aware characteristics:
\begin{enumerate}
\item	The in-vehicle features related to the driver distraction such as hand moment, gaze;
\item	The vehicle features such as vehicle speed, manoeuvres;

\item The environmental features such as pedestrians, vehicles, and weather.
\end{enumerate}
The association and interdependencies between a pair of distractions need to be tested. We applied a regression to the prediction of driver distraction divided into severity levels. Here, we further tested the relationship between distractions by classifying distraction as either in-vehicle, context-aware, or environmental distraction.

\subsubsection {Driver Distraction}
\begin{figure}[H]
	\centering
	\includegraphics[width=0.8\linewidth]{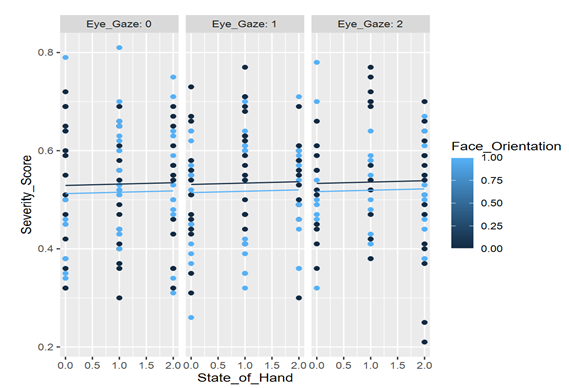}
	\caption{In-Vehicle State of Hand, Eye Gaze, and Face Orientation}
	\label{fig:fig5}
\end{figure}

if they are meant to be capitalised they will go automatically, but if we change the forum then we have to do it manually every time
Driver distraction features consist of state of hand, face orientation, and eye gaze. Figure \ref{fig:fig5} shows driver distraction, showing a strong relationship between eye gaze on the road (eye gaze 0), off-road face orientation and a single hand on the wheel, and having a high severity level of distraction leading to dangerous driving. Eyes closed, the orientation of the face on the road, and double hands-on wheel also significantly affect the severity score.

Table \ref{tab7} presents the distractions in the vehicle with respect to the prediction of the severity score. Based on the $P$ value of 0.758, the probability of the state of hands to predict the severity of distraction is low. The intercept of 0.529134, which is highly significant, suggests a relationship between the state of hands, face orientation, and eye gaze. The statistical predictors use the t-statistics and P-values of each distraction. The lower P-value of 0.249 for face orientation shows that this is a highly significant predictor of the severity score. The coefficient of determination is 0.005668.

\begin{table}[ht!]
\centering
		\caption{Human Distractions Regression Analysis}
	\label{tab7}
	\begin{tabular}{|l|l|l|l|l|}
		\hline
		\textbf{Intercepts} & \textbf{Estimated} & \textbf{Std   Error} & \textbf{T-Value} & \textbf{P Value} \\ \hline
		State   of Hand     & 0.015526           & 0.015526                  & 34.080           & 0.758            \\ \hline
		Face   Orientation  & 0.1-0.25           & 0.014414                  & -1.156           & 0.249            \\ \hline
		Eye   Gaze          & 0.002045           & 0.008981                  & 0.228            & 0.820            \\ \hline
		Intercept           & 0.529134           & 0.015526                  & 34.080           & \textless{}2e-16 \\ \hline
	\end{tabular}
\end{table}

\subsubsection {Environmental Distractions}
\begin{figure}[ht!]
	\centering
	\includegraphics[width=0.8\linewidth]{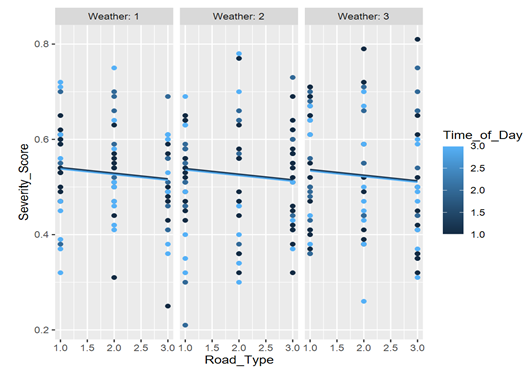}
	\caption{Road Type, Time of Day, and Weather}
	\label{fig:fig6}
\end{figure}

Figure \ref{fig:fig6} shows that dry weather, dual carriageway, and a bright day produced the highest dangerous severity score, while rainy weather, double carriageway, and night produced a slightly more dangerous situation. Snowy conditions on the highway and at night had the highest degree of influence on the severity score. 

The results in Table \ref{tab8} show an environmental distraction about the prediction of the severity score. The low $P$-value of 0.175 for the road type shows that the road type has a highly significant impact on the prediction. The environment intercept of 0.556982 showed a significant association between the severity score and the type of distraction road of the result, the time of day and the weather. However, the lowest average P-value of 0.5897 was the lowest compared to the other distraction classifications. The least residual standard error of 0.1165, which is the smallest of all residual standard errors, indicates that this model best fits the data.
\begin{table}[ht!]
\centering
		\caption{Environment Regression Analysis}
	\label{tab8}
	\begin{tabular}{|l|l|l|l|l|}
		\hline
		\textbf{Intercepts} & \textbf{Estimated} & \textbf{Std  Error} & \textbf{T-Value} & \textbf{P-Value} \\ \hline
		Road   Type         & -0.011838          & 0.008702                 & -1.360           & 0.175            \\ \hline
		Time of Day         & -0.011848          & 0.008579                 & -0.215           & 0.830            \\ \hline
		Weather             & -0.002086          & 0.009035                 & -0.231           & 0.818            \\ \hline
		Intercept           & 0.556982           & 0.031870                 & 17.477           & \textless{}2e-16 \\ \hline
	\end{tabular}
\end{table}

\begin{figure}[H]
	\centering
	\includegraphics[width=0.8\linewidth]{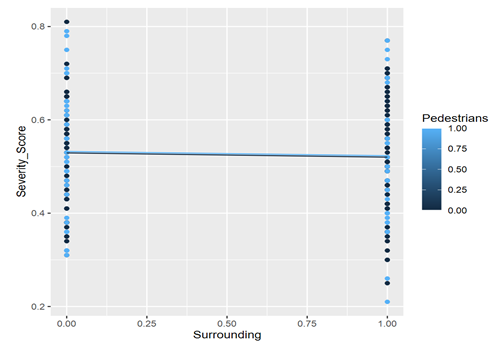}
	\caption{Pedestrian and Surrounding (Vehicle Presence)}
	\label{fig:fig7}
\end{figure}

Figure \ref{fig:fig7} shows five instances of vehicle presence and pedestrian presence that result in a hazardous distraction classification, which means that if there are vehicles or pedestrians present in the surroundings, the chances of distraction by the driver are significant.

Table \ref{tab9} presents the in-vehicle distractions to the prediction of the severity score. Based on the P-value of 0.532, the probability that the surroundings influence the prediction is low. The intercept of 0.529157, which is significant, suggests a relationship between the surroundings and pedestrians. The statistical predictors use the t-statistics and P-values of each distraction. The lower P-value of 0.532 for the surroundings (vehicle presence) is a highly significant predictor of the severity score. However, the P-value of 0.830 for pedestrians suggests that there is no association between pedestrians and the severity score.

\begin{table}[H]
\centering
	\caption{External Distractions Regression Analysis}
	\label{tab9}
	\begin{tabular}{|l|l|l|l|l|}
		\hline
		\textbf{Intercepts} & \textbf{Estimated} & \textbf{Std   Error} & \textbf{T-Value} & \textbf{P-Value} \\ \hline
		Surrounding         & -0.009065          & 0.014502                  & -0.625           & 0.532            \\ \hline
		Pedestrians         & 0.003121           & 0.014487                  & 0.215            & 0.830            \\ \hline
		Intercept           & 0.529157           & 0.013145                  & 40.255           & \textless{}2e-16 \\ \hline
	\end{tabular}
\end{table}

\begin{figure}[ht!]
	\centering
	\includegraphics[width=0.8\linewidth]{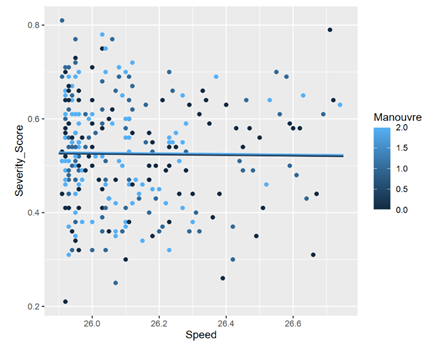}
	\caption{Vehicle, Speed and Manoeuvre.}
	\label{fig:fig8}
\end{figure}

Vehicle distractions include manoeuvres and speed. Figure \ref{fig:fig8} shows the distraction within the speed range of 23 mph to 26.2 mph due to the high frequency of speed manoeuvres. There are a few outliers with very high severity and very high levels of danger.

The results in Table \ref{tab10} show the influence of vehicle distractions on the prediction of a severity score. Based on the p-value of 0.855, the probability that speed influences the prediction is low because the driver stays within the speed limit. However, during manoeuvres, there is a higher degree of significance. The intercept of 0.695812, which is highly significant, suggests a strong relationship between speed and manoeuvre. The statistical predictors use the t-statistics and P-values of each distraction. The lower P-value of 0.815 for manoeuvres shows that this is a highly significant predictor of the severity score.

\begin{table}[ht!]
\centering 
	\caption{Vehicle Distractions Regression Analysis}
	\label{tab10}
	\begin{tabular}{|l|l|l|l|l|}
		\hline
		\textbf{Intercepts} & \textbf{Estimated} & \textbf{Std   Error} & \textbf{T-Value} & \textbf{P-Value} \\ \hline
		Speed               & -0.009065          & 0.035983                  & -0.183           & 0.855            \\ \hline
		Manoeuvre           & 0.002050           & 0.008758                  & 0.234            & 0.815            \\ \hline
		Intercept           & 0.695812           & 0.941042                  & 0.739            & 0.460            \\ \hline
	\end{tabular}
\end{table}

\begin{figure}[ht!]
	\centering
	\includegraphics[width=0.8\linewidth]{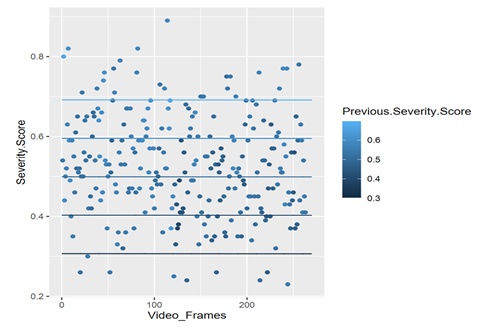}
	\caption{Severity Score, Previous Severity Score and Video Frames}
	\label{fig:fig9}
\end{figure}

In this case, the driver is tested with the previous severity score and the predicted actual severity score of the following video frame; in this experiment, the driver's overall performance is tested throughout the drive, whereby there are a total of 262 frames, which is the equivalent of approximately 11 seconds. In Figure \ref{fig:fig9}, we can assume that the driver has maintained a primarily constant careless driving behaviour. Additionally, the regression analysis results in a strong correlation between the previous severity score and the current severity score.  

The results presented in Table \ref{tab11} show the influence of distractions in the vehicle on the prediction of the severity score. Based on the P-value with a lower value of 0.990, the state-of-hand probability predicts that the score is low. The intercept of 1.828e, which is highly significant, suggests a relationship between the sequence of video frames and the previous severity score.

\begin{table}[ht!]
\centering
		\caption{Severity Score Regression Analysis}
	\label{tab11}
  \begin{adjustbox}{ width = 0.99\textwidth}
	\begin{tabular}{|l|l|l|l|l|}
		\hline
		\textbf{Intercepts}       & \textbf{Estimated} & \textbf{Std   Error} & \textbf{T-Value} & \textbf{P-Value} \\ \hline
		Video   Frames            & -1.329e-06         & 1.024e-04                 & -0.013           & 0.990            \\ \hline
		Previous   Severity Score & 9.618e-01          & 9.618e-01                 & 5.812            & 1.8e-08          \\ \hline
		Intercept                 & 1.828e-02          & 1.828e-02                 & 0.196            & 0.845            \\ \hline
	\end{tabular}
 \end{adjustbox}
\end{table}

\subsection{ML MODEL} 
We implemented different ML models, such as discriminant, N\"ive Bayes, SVM, K-Means Nearest Neighbour (KNN), and Ensemble ML. To better evaluate performance, the comparison of the classification results is shown in Table \ref{tab12}.

\begin{table}
\centering
	\caption{Classification performances for the KNN, Discriminant, Na\"ive Bayes, SVM and Ensemble models.}
	\label{tab12}
	\begin{tabular}{|l|l|l|l|}
        
		\hline
		\textbf{Model}    & \textbf{Acc. \%} & \textbf{Speed} & \textbf{T-Time} \\ \hline
		 Fine   KNN               & 79.1                   & 2700                      & 4.4574                   \\ 
		      Medium   KNN             & 78.3                   & 2500                      & 3.5617                   \\  
		 KNN Coarse             & 59.3                   & 2500                      & 4.4974                   \\  
		 KNN   Cosine             & 80.6                   & 2600                      & 4.368                    \\  
		 KNN   Cubic              & 76.4                   & 2000                      & 4.239                    \\  
		 KNN   Weighted KNN       & 80.6                   & 2500                      & 3.975                    \\ \hline
		 Linear   Discriminant    & 90.9                   & 2700                      & 3.5265                   \\ 
		 Quadratic   Discriminant & 82.9                   & 2500                      & 5.2346                   \\ \hline
		 Gaussian   Na\"ive Bayes   & 93.2                   & 3000                      & 5.0814                   \\ 
		 Kernel   Na\"ive Bayes     & 90.1                   & 1500                      & 5.9402                   \\ \hline
	            Linear   SVM             & 92.0                   & 2400                      & 4.9151                   \\ \hline
		 Quadratic   SVM          & 92.4                   & 2300                      & 4.8007                   \\  
		 Cubic   SVM              & 92.4                   & 2300                      & 4.6915                   \\  
		 Fine   Gaussian SVM      & 58.6                   & 2200                      & 5.7229                   \\  
		 Medium   Gaussian SVM    & 85.2                   & 2100                      & 5.5983                   \\  
		 Coarse   Gaussian SVM    & 77.2                   & 2300                      & 5.4722                   \\ \hline
		 Boosted Trees            & 58.6                   & 3600                      & 4.5331                   \\ \hline
		 Bagged   Trees           & 96.2                   & 1000                      & 6.3019                   \\  
		 Subspace   Discriminant  & 92.4                   & 780                       & 6.8675                   \\  
	 Subspace   KNN           & 79.8                   & 600                       & 6.7319                   \\  
		 RUSBoosted   Trees       & 74.5                   & 2900                      & 4.6438                   \\ \hline
	\end{tabular}
\end{table}

 In Figure \ref{fig:fig10}, the first two diagonal cells show the percentage of correct classification by the trained network. For example, 142 frames are correctly classified as careless. This corresponds to 99\% of the 262 frames. Similarly, 80 cases are correctly classified as dangerous, which corresponds to 96\% of all edges. Three dangerous and three safe instances are incorrectly classified, corresponding to 12\% of all 264 frames in the data. Similarly, one of the careless structures is incorrectly classified, corresponding to 1\% of all data. Out of 148 careless predictions, 99\% are correct and 1\% are wrong. Of the 80 dangerous predictions, 96\% are correct and 4\% are wrong. Of 35 safe cases, 92\% are correctly predicted as safe, and 8\% are correctly predicted as false.

\begin{figure}[ht!]
    \centering
    \begin{tabular}{cr|ccc|}
       & \multicolumn{1}{c}{}  & \multicolumn{3}{c}{\textbf{Predicted}}  \\
           & \multicolumn{1}{c}{} & \multicolumn{1}{c}{Careless} & Dangerous & \multicolumn{1}{c}{Safe} \\
           \cline{3-5}
        \multirow{3}{*}{\textbf{Actual}} & Careless & 99 & 4 & 8 \\
                                & Dangerous & 0 & 96 & 0 \\
                                & Safe & 1 & 0 & 92 \\
                                \cline{3-5}\\
         \multicolumn{2}{c}{False Discovery Rate} &1 &4 &\multicolumn{1}{c}{8}
    \end{tabular}
    \caption{Confusion matrix, showing the predicted and actual safe, careless, and dangerous driving.}
    \label{fig:fig10}
\end{figure}

 \subsubsection{Time Complexity}
\begin{figure}[ht!]
\centering
\begin{subfigure}{0.49\textwidth}
    \includegraphics[width=\textwidth,trim={0 0 0 0.6cm},clip]{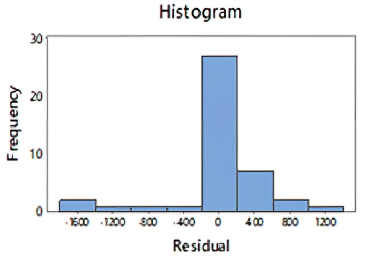}
    \caption{Histogram of Residuals}
    \label{fig:histrogramresidual}
\end{subfigure}
\begin{subfigure}{0.49\textwidth}
    \includegraphics[width=\textwidth,trim={0 0 0 0.6cm},clip]{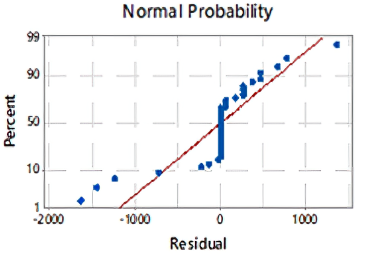}
    \caption{Linear discriminant}
    \label{fig:slopeb}
\end{subfigure}

\begin{subfigure}{0.5\textwidth}
    \includegraphics[width=\textwidth,trim={0 0 0 0.6cm},clip]{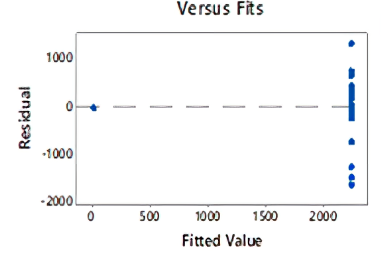}
    \caption{Approximate Variance Inflation Factor}
    \label{fig:versusc}
\end{subfigure}

\caption{Residual Plots for Prediction Speed (obs/sec) vs Training Time (sec).}
\label{fig:fig12}
\end{figure}

Safety in intelligent transportation systems (ITS) is critical, and having a fast ML model to make an efficient decision is crucial for the safety of road users. Figure
\ref{fig:fig12} refers to the amount of time it takes to train a model, make predictions, or perform other operations related to the learning process. The histogram of residual values  displays the frequency distribution of the residuals in graph (a) indicates the distance between the observed prediction time from the mean of each classifier's total time for training. The significant residual value between -200 and +300 shows an optimal configuration for the proposed framework when employing ML variants. shows that the fitted line's intercept and slope values are projections for the distribution's position and residual parameters, respectively. 
Linear discriminant in (b) depicts the shortest training time of 3.5265. This means that this algorithm is able to train a model to classify data points into different categories more quickly than the other algorithms being compared. The y-axis of the graph shows the percentage accuracy of the different algorithms. The sample variance, which is a measure of the spread of the data, is being used to approximate the accuracy, prediction time, and training time obtained using several ML algorithms. 
The residual values in (c) y-axis show that the prediction was exceedingly low. Fitted values (refer to the x-axis) show that the prediction was significantly accurate; 0 on the y-axis indicates a 100\% correct positive rate.

\subsubsection{Kruskal-Wallis Results}

The Kruskal-Wallis rank test was used to evaluate the algorithms and the results were presented in \ref{tab13}. The authors observed that the ensemble learning method with the Bagged model had the highest mean rank of 21 compared to the other variants of that model and other state-of-the-art ML algorithms. This suggests that the Bagged model's complex fitness function helped to extract rich feature vectors for classification. On the other hand, the Boosted Trees model had the highest mean rank of 21 compared to the other variants of that model and other state-of-the-art ML algorithms. However, the authors noted that this model's linear fitness function helps to extract poor feature vectors for classification. Additionally, the authors found that the linear discriminant variant had the lowest mean rank of 1 in training time and achieved 90\% accuracy. This algorithm used linear functions to evaluate the previous severity score and the expected total severity score of the following video frame. Finally, the authors noted that Gaussian N\"ive Bayes was the second-best algorithm in terms of performance compared to others, except for bounded trees. This algorithm had an average mean rank of 20, achieved 93.2\% accuracy, and had a z-score of 1.49. Overall, the results suggest that the Bagged model and Gaussian N\"ive Bayes performed well in this experiment, while the Boosted Trees model may not be the best option for feature extraction in classification tasks.

\begin{table}
\centering
	\caption{Kruskal- Wallis ranks for the ML Algorithms to confirm Accuracy ($\alpha$), Training Time ($\tau_T$), and Prediction Time ($\tau_P$), Z-Score ($\zeta$), and the median value ($\mu$).}
	\label{tab13}
 
\begin{tabular}{|c|ccccc|}
		\hline
		\multirow{2}{*}{\textbf{Model}} & \multicolumn{5}{c|}{\textbf{Kruskal-Wallis Average Ranks}}   \\ 
		                 & $\boldsymbol{\mu}$ & $\boldsymbol{\alpha}$ & $\boldsymbol{\tau_P}$ & $\boldsymbol{\tau_T}$ &        $\boldsymbol{\zeta}$                  \\ \hline
		Bagged Trees          & 96.2  & 21       & 3          & 19     & 1.65             \\ \hline
		Boosted Trees         & 58.6  & 1.5     & 21        & 8      & -1.57            \\ \hline
		Coarse Gaussian SVM   & 77.2  & 6       & 9         & 15     & -0.83           \\ \hline
		Cubic SVM             & 92.4  & 18      & 9         & 10     & 1.16            \\ \hline
		Fine Gaussian SVM     & 58.6  & 1.5     & 7         & 17     & -1.57            \\ \hline
		Fine KNN              & 79.1  & 8       & 17.5      & 6      & -0.5              \\ \hline
		Gaussian {N\"ive} Bayes  & 93.2  & 20      & 20        & 13     & 1.49              \\ \hline
		Kernel {N\"ive} Bayes    & 90.1  & 14      & 4         & 18     & 0.5            \\ \hline
		KNN Coarse            & 59.3  & 3       & 13.5      & 7      & -1.32         \\ \hline
		KNN Cosine            & 80.6  & 10.5    & 16        & 5      & -0.08         \\ \hline
		KNN Cubic             & 76.4  & 5       & 5         & 4      & -0.99         \\ \hline
		KNN Weighted KNN      & 80.6  & 10.5    & 13.5      & 3      & -0.08         \\ \hline
		Linear Discriminant   & 90.9  & 15      & 17.5      & 1      & 0.66          \\ \hline
		Linear SVM            & 92    & 16      & 11        & 12     & 0.83          \\ \hline
		Medium Gaussian SVM   & 85.2  & 13      & 6         & 16     & 0.33          \\ \hline
		Medium KNN            & 78.3  & 7       & 13.5      & 2      & -0.66         \\ \hline
		Quadratic Discriminant & 82.9  & 12      & 13.5      & 14     & 0.17          \\ \hline
		Quadratic SVM         & 92.4  & 18      & 9         & 11     & 1.16          \\ \hline
		RUSBoosted Trees      & 74.5  & 4       & 19        & 9      & -1.16         \\ \hline
		Subspace Discriminant & 92.4  & 18      & 2         & 21     & 1.16          \\ \hline
		Subspace KNN          & 79.8  & 9       & 1         & 20     & -0.33         \\ \hline
	\end{tabular}
\end{table}

\subsubsection{Comparison of Models}

\begin{figure}[ht!]
\begin{tikzpicture}
    \begin{axis}[
        width  = \textwidth,
        height = 6cm,
        cycle list name=fc,
        bar width=28pt,
        ymajorgrids = true,
        ylabel = {\textbf{Accuracy (\%)}},
        xtick = {1,2,3,4},
        xticklabels = {\cite{b70},\cite{b71},\cite{b72},MDDRA},      
        enlarge x limits=0.25,
        ymin=80,
        nodes near coords,
    ] 
\addplot+[ybar]coordinates {(1,95.87)};       
\addplot+[ybar]coordinates {(2,89.9)};
\addplot+[ybar]coordinates {(3,97.21)}; 
\addplot+[ybar]coordinates {(4,96.21)}; 
\end{axis}
\end{tikzpicture}
\caption{Accuracy in \% Model comparison}
	\label{fig:fig13}
\end{figure}
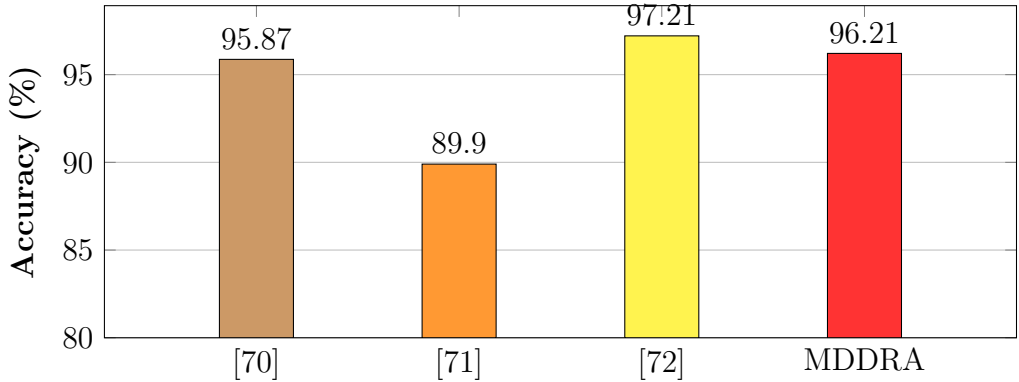

The graph in Figure \ref{fig:fig13} compares the accuracy value of the proposed MDDRA, the work of Mengtao Zhu et al. \cite{b70}, the work proposed by Yanli Ma et al. \cite{b71}, and the work of Tianchi Liu et al. \cite{b72}. It can be seen that the proposed model has outperformed the current state-of-the-art in multiclass distraction prediction. Moreover, the model has achieved an accuracy of 96.21\%, while the current state-of-the-art claims an accuracy of 95.87\%, which is lower than our proposed methodology. Although Tianchi Liu et al. \cite{b72} have achieved slightly higher accuracy, they have worked on a binary classification problem. The multiclass classification is a more complex task than a simple binary classification model, the state-of-the-art model with excellent results in more than eight classes. Furthermore, the proposed model has provided fast results as high as 3600 observations per second, making the proposed model accurate but robust in terms of speed.

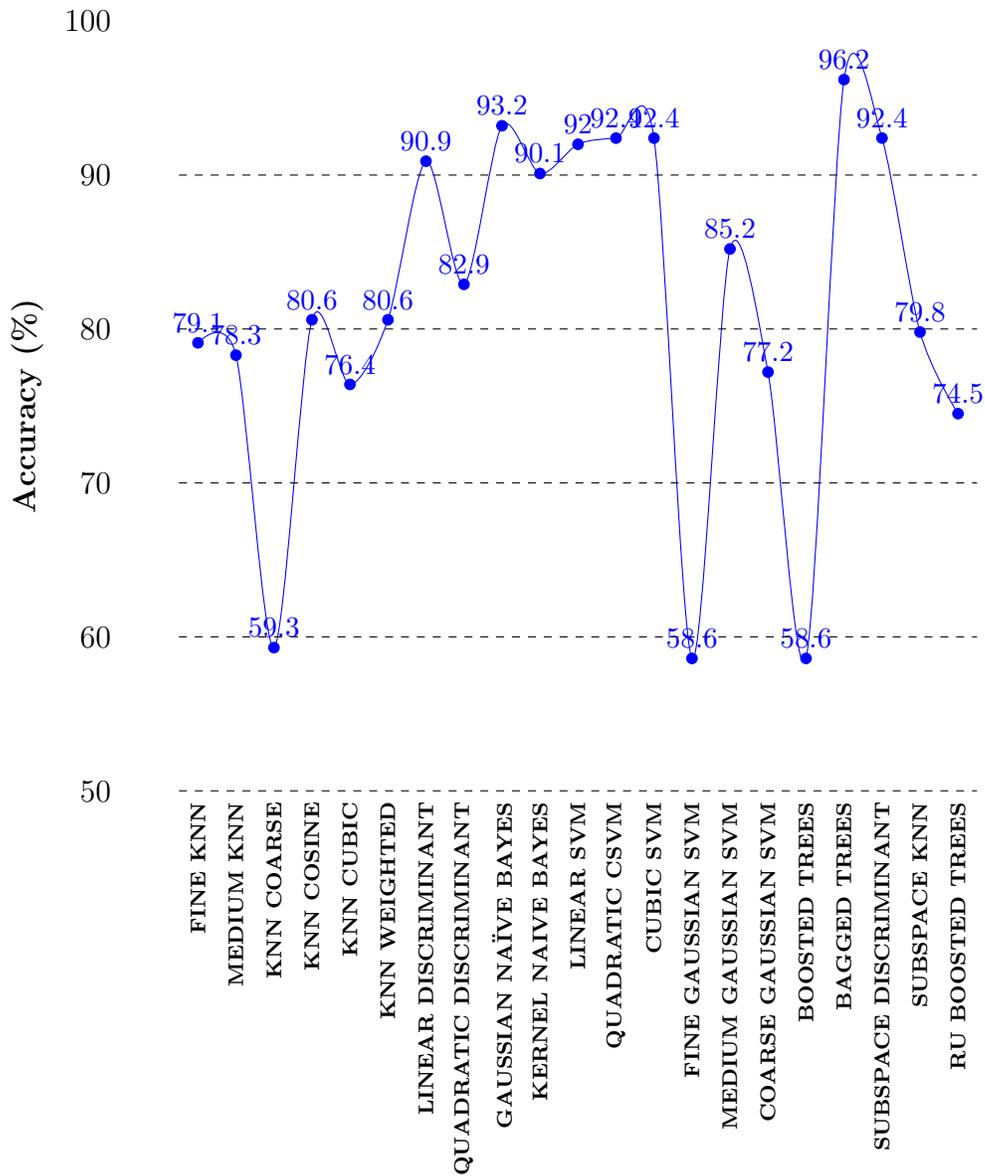
\begin{figure}[ht!]
\centering
\begin{tikzpicture}

\begin{axis}[
  width=\textwidth,
  xlabel style={yshift=-125pt},
  ylabel=\textbf{Accuracy (\%)},
  ymin=50,
  ymax=100,
    ytick={50,60,70,80,90,100},
  xtick=data,
  xticklabels={\scriptsize\textbf{FINE KNN}, \scriptsize\textbf{MEDIUM KNN}, \scriptsize\textbf{KNN COARSE}, \scriptsize\textbf{KNN COSINE},\scriptsize\textbf{KNN CUBIC},\scriptsize\textbf{KNN WEIGHTED},\scriptsize\textbf{LINEAR DISCRIMINANT},\scriptsize\textbf{QUADRATIC DISCRIMINANT},\scriptsize\textbf{GAUSSIAN NA\"IVE BAYES},\scriptsize\textbf{KERNEL NAIVE BAYES},\scriptsize\textbf{LINEAR SVM},\scriptsize\textbf{QUADRATIC CSVM},\scriptsize\textbf{CUBIC SVM},\scriptsize\textbf{FINE GAUSSIAN SVM},\scriptsize\textbf{MEDIUM GAUSSIAN SVM},\scriptsize\textbf{COARSE GAUSSIAN SVM},\scriptsize\textbf{BOOSTED TREES},\scriptsize\textbf{BAGGED TREES},\scriptsize\textbf{SUBSPACE DISCRIMINANT},\scriptsize\textbf{SUBSPACE KNN},\scriptsize\textbf{RU BOOSTED TREES}},
  xticklabel style={rotate=90,anchor=near xticklabel},
  nodes near coords,
  nodes near coords align={vertical},
  nodes near coords style={font=\small},
  axis line style={draw=none},
  tick style={draw=none},
  legend style={at={(0.83,1.2)},anchor=north},
  legend cell align={left},
  ]
  
  \addplot[smooth,mark=*,blue] coordinates {
  (1,79.1)
  (2,78.3)
  (3,59.3)
  (4,80.6)
  (5,76.4)
  (6,80.6)
  (7,90.9)
  (8,82.9)
  (9,93.2)
  (10,90.1)
  (11,92)
  (12,92.4)
  (13, 92.4)
  (14,58.6)
  (15,85.2)
  (16,77.2)
  (17,58.6)
  (18,96.2)
  (19,92.4)
  (20,79.8)
  (21,74.5)
  };
\draw[dashed] (axis cs:0.5,90) -- (axis cs:21.5,90);
\draw[dashed] (axis cs:0.5,80) -- (axis cs:21.5,80);
\draw[dashed] (axis cs:0.5,70) -- (axis cs:21.5,70);
\draw[dashed] (axis cs:0.5,60) -- (axis cs:21.5,60);
\draw[dashed] (axis cs:0.5,50) -- (axis cs:21.5,50);
\end{axis}
\end{tikzpicture}
\caption{Accuracy across multiple ML models}
\label{fig:fig14}

\end{figure}

This paper deals with the problem by employing machine learning. The authors have proposed a novel and robust 
 MDDRA model. The model has tackled the driver with almost possible variants, such as the current state of the hand, which means whether the driver uses double hands, single hands, or no hands. Similarly, the type of road on which the vehicle is running, the face orientation on the road or off-road, whether it is a day or night, the eye gaze of the driver, if the weather is dry, rainy or snowy, what is a current manoeuvre, the surrounding vehicles, speed of the vehicle, speed of the surrounding vehicle, and pedestrians. The suggested model, MDDRA, considers vehicle, driver, and environmental data that occur during a journey to categorise drivers into a risk matrix such as safe, careless, and dangerous. 
The proposed model offers the flexibility to adjust parameters and weights to consider each event at a specific severity level. Real-world data was collected using the Field Operation Test (TeleFOT), which consisted of drivers using the same routes in the East Midlands, UK. The results have great potential to reduce road accidents caused by distracted drivers. We have also tested the correlation of driver distraction (in-vehicle, vehicle, and environment distractions) on severity classification against the continuous driver distraction severity score. Furthermore, we have applied several machine learning techniques to classify and predict driver distraction according to severity levels to help transition from driver to vehicle. 
We implemented different ML models such as Discriminant, N\"ive Bayes, Support Vector Machine (SVM), and K-Means Nearest Neighbour (KNN) Ensemble ML for classification. Figure \ref{fig:fig14} shows the comparison of accuracy by applying these models. It can be seen that the Bagged Trees-based Ensemble model has provided the highest accuracy of 96.2\% for classification, while fine Gaussian SVM and Boosted Trees-based ensemble methods have resulted in the lowest accuracy of 58.6\% for the classification task. The comparison of various ML models is shown in Figure \ref{fig:fig14}.

\section{CONCLUSIONS}\label{sec:con}
For a robust false proof alert system, the precise classification of driving behaviour is needed. However, to the best of our knowledge, the current work lacks complexity, rigidity, a synthesised dataset, more focus on a particular side of perspective (vehicle, driver, or environment), false positive classes, and low accuracy.  This paper aimed to provide a novel MDDRA model that considers vehicle, driver, and environmental data during a trip to categorise the driver in a risk matrix as safe, careless, or dangerous. The MDDRA model offers flexibility in adjusting the parameters and weights to consider each event on a specific severity level. Real-world data was collected using the Field Operation Test (TeleFOT), consisting of drivers using the same routes in the East Midlands, United Kingdom (UK). The results showed that it is possible to reduce road accidents caused by distracted drivers. We also tested the correlation between distraction (driver, vehicle, and environment) and classification severity based on a continuous distraction severity score.

Furthermore, we applied machine learning techniques to classify and predict driver distraction according to severity levels to aid the transition of control from the driver to the vehicle (vehicle takeover) when a situation is deemed risky. The experimental results obtained using various ML algorithms have shown better results than those of the baseline and the previous literature. The algorithm with the best performance was Ensemble Bagged Trees, which gave an accuracy of 96.2 \%.
However, this approach's limitation is that deep learning will produce better results regarding speed performance than an ML technique.  The result of the vehicle regression analysis had a higher degree of correlation and was highly significant. The MDDRA model can be adjusted to fit any distraction risk assessment considering the driver, vehicle, and environmental contexts. When assigning weights to pedestrians on the road, we did not consider accidents or vehicles that hit pedestrians. However, the results of the regression show that vehicle distraction constitutes a higher level of significance. Factors such as sample size and data spread may have influenced the regression analysis's P-value results. Confidence intervals around the sample statistics would yield a better result than P-values alone. In addition, adopting deep learning Convolutional Neural Network-Long Short-Term Memory (CNN-DBN-LSTM) techniques in detecting and classifying multiclass driver distraction would yield more effective and efficient results. Finally, considering the accuracy over time complexity, the best ML model adopted is the Bagged Trees.

\bibliographystyle{elsarticle-harv}

\end{document}